\documentclass[letterpaper]{article}
\usepackage{aaai}
\usepackage{fixbib}
\usepackage{times}
\usepackage{helvet}
\usepackage{courier}
\usepackage{booktabs}

\usepackage[utf8]{inputenc} 
\usepackage[T1]{fontenc}    
\usepackage{hyperref}       
\usepackage{url}            
\usepackage{booktabs}       
\usepackage{amsfonts}       
\usepackage{nicefrac}       
\usepackage{microtype}      
\usepackage{color}
\usepackage{amssymb}
\usepackage{amsmath}
\usepackage{graphicx}
\usepackage{algorithm}
\usepackage[noend]{algpseudocode}
\usepackage{amsthm}
\usepackage{subfig}
\usepackage{tablefootnote}
\usepackage{threeparttable}
\usepackage{multirow}

\theoremstyle{definition}
\newtheorem*{remark}{Remark}

\allowdisplaybreaks[4]
\newcommand{\tabincell}[2]{
\begin{tabular}{@{}#1@{}}#2\end{tabular}
}

\usepackage{array}
\newcommand{\PreserveBackslash}[1]{\let\temp=\\#1\let\\=\temp}
\newcolumntype{C}[1]{>{\PreserveBackslash\centering}p{#1}}
\newcolumntype{R}[1]{>{\PreserveBackslash\raggedleft}p{#1}}
\newcolumntype{L}[1]{>{\PreserveBackslash\raggedright}p{#1}}

\begin{document}

\title{Accelerating Primal Solution Findings for Mixed Integer Programs\\ Based on Solution Prediction }
\author{Jian-Ya Ding\textsuperscript{1}, Chao Zhang\textsuperscript{1}, Lei Shen\textsuperscript{1}, Shengyin Li\textsuperscript{1}, Bing Wang\textsuperscript{1}, Yinghui Xu\textsuperscript{1}, Le Song\textsuperscript{2}\\
\textsuperscript{1}: Artificial Intelligence Department, Zhejiang Cainiao Supply Chain Management Co., Ltd\\
\textsuperscript{2}: Computational Science and Engineering College of Computing, Georgia Institute of Technology\\
}


\maketitle
\begin{abstract}
Mixed Integer Programming (MIP) is one of the most widely used modeling techniques for combinatorial optimization problems. In many applications, a similar MIP model is solved on a regular basis, maintaining remarkable similarities in model structures and solution appearances but differing in formulation coefficients. This offers the opportunity for machine learning methods to explore the correlations between model structures and the resulting solution values. To address this issue, we propose to represent an MIP instance using a tripartite graph, based on which a Graph Convolutional Network (GCN) is constructed to predict solution values for binary variables. The predicted solutions are used to generate a local branching type cut which can be either treated as a global (invalid) inequality in the formulation resulting in a heuristic approach to solve the MIP, or as a root branching rule resulting in an exact approach. Computational evaluations on 8 distinct types of MIP problems show that the proposed framework improves the primal solution finding performance significantly on a state-of-the-art open-source MIP solver.
 \end{abstract}

\section{Introduction}

\emph{Mixed Integer Programming} (MIP) is widely used to solve \emph{combinatorial optimization} (CO) problems in the field of \emph{Operations Research} (OR). The existence of integer variables endows MIP formulations with the ability to capture the discrete nature of many real-world decisions.  Applications of MIP include production scheduling \cite{chen2010integrated}, vehicle routing \cite{laporte2009fifty}, facility location \cite{farahani2009facility}, airline crew scheduling \cite{gopalakrishnan2005airline}, to mention only a few. In many real-world settings, homogeneous MIP instances with similar scales and combinatorial structures are optimized repeatedly but treated as completely new tasks. These MIP models share remarkable similarities in model structures and solution appearances, which motivates us to integrate {\it Machine Learning} (ML) methods to explore correlations between an MIP model's structure and its solution values to improve the solver's performance.

Identifying correlations between problem structures and solution values is not new, and is widely used as guidelines for heuristics design for CO problems. These heuristic methods are usually human-designed priority rules to guide the search directions to more promising regions in solution space. For example, the nearest neighbor algorithm \cite{cook2011pursuit} for the traveling salesman problem (TSP) constructs a heuristic solution by choosing the nearest unvisited node as the salesman's next move, based on the observation that two distantly distributed nodes are unlikely to appear consecutively in the optimal route. Similar examples include the shortest processing time first heuristic for flow shop scheduling \cite{pinedo2012scheduling}, the saving heuristic for vehicle routing \cite{clarke1964scheduling}, the first fit algorithm for bin packing \cite{dosa2013first}, among many others.  A major drawback of heuristics design using problem-specific knowledge is the lack of generalization to other problems, where new domain knowledge has to be re-identified.

MIP models can describe CO problems of various types using a standard formulation strategy $
z = \min_{\boldsymbol{Ax\le b}, \boldsymbol{x}\in\mathcal{X}}~\boldsymbol{c}^T\boldsymbol{x}$, differing only in model coefficients $\boldsymbol{A}, \boldsymbol{b}$ and $\boldsymbol{c}$ and integrality constraints $\mathcal{X}$. This makes it possible to explore connections between problem structures and the resulting solution values without prior domain knowledge. Predicting solution values of general integer variables in MIP is a difficult task. Notice that most MIP models are binary variables intensive\footnote{Take the benchmark set of MIPLIB 2017 \cite{miplib2017} as an example, among all 240 MIP benchmark instances, 164 of them are Binary Integer Linear Programming (BILP) problems, and 44 out of the 76 remainings are imbalanced in the sense that binary variables account for more than 90\% of all integer variables.}, a natural way to explore hidden information in the model is to treat solution value prediction of binary variables as binary classification tasks. Major challenges in solution prediction lie in the implicit correlations among decision variables, since a feasible solution $\boldsymbol{x}$ is restricted by constraints in MIP, i.e., $\boldsymbol{Ax\le b}$. Rather than predicting each decision variable value isolatedly, we propose a tripartite graph representation of MIP and use graph embedding to capture connections among the variables. Note that none of the two variable nodes are directly linked in the trigraph, but can be neighbors of distance $2$ if they appear in the same constraint in the MIP formulation. Correlations among variables are reflected in embeddings of the trigraph where each vertex maintains aggregate feature information from its neighbors.

Incorporating solution prediction results in MIP solving process is not trivial. Fixing a single false-predicted decision variable can sometimes lead to the infeasibility of the entire problem. Instead of utilizing the predicted solutions directly, we identify predictable decision variables and use this information to guide the Branch and Bound (B\&B) tree search to focus on unpredictable ones to accelerate convergence. This is achieved by a novel labeling mechanism on the training instances, where a sequence of feasible solutions is generated by an iterated proximity search method. \emph{Stable} decision variables, of which the value remain unchanged across these solutions, are recorded. It is noticeable that although obtaining optimal solutions can be a difficult task, the stable variables can be viewed as an easy-to-predict part that reflects the MIP's local optimality structure. This labeling mechanism is inspiring especially for difficult MIP instances when solving them to optimality is almost impossible.

The overall framework of solution prediction based MIP solving can be summarized as follows:

\vspace{1mm}
\noindent\textbf{Training data generation}:  For a certain type of CO problem, generate a set of $p$ MIP instances $\mathcal{I}= \{I_1, \ldots, I_p\}$ of similar scale from the same distribution $\mathbb{D}$. For each $I_i\in \mathcal{I}$, collect variable features, constraint features, and edge features, and use the iterated proximity search method to generate solution labels for each binary variable in $I_i$.\\

\vspace{-3mm}
\noindent\textbf{GCN model training}: For each $I_i\in \mathcal{I}$, generate a tripartite graph from its MIP formulation. Train a Graph Convolutional Network (GCN)  for binary variable solution prediction based on the collected features, labels and trigraphs.\\

\vspace{-3mm}
\noindent\textbf{Application of solution prediction}: For a new MIP instance $I$ from $\mathbb{D}$, collect features, build the trigraph and use the GCN model to make solution value predictions, based on which an initial local branching cut heuristic (or a root branching exact approach) is applied to solve $I$.

\section{Related work}
With similar motivation, there are some recent attempts that consider the integration of ML and OR for solving CO problems. \cite{dai2017learning} combined reinforcement learning and graph embedding to learn greedy construction heuristics for several optimization problems over graphs. \cite{li2018combinatorial} trained a graph convolutional network to estimate the likelihood that a vertex in a graph appears in the optimal solution. \cite{selsam2018learning} proposed a message passing neural network named NeuroSAT to solve SAT problems via a supervised learning framework. \cite{vinyals2015pointer},  \cite{kool2018attention2,kool2018attention} and \cite{nazari2018reinforcement} trained Pointer Networks (or its variants) with recurrent neural network (RNN) decoder to solve permutation related optimization problems over graphs, such as Traveling Salesman Problem (TSP) and Vehicle Routing Problem (VRP). Different from their settings, our solution prediction framework does not restrict to certain graph-based problems but can adapt to a variety of CO problems using a standard MIP formulation.

Quite related to our work, there is an increasing concern in using ML techniques to enhance MIP solving performance. \cite{alvarez2017machine}, \cite{marcos2016online}, \cite{khalil2016learning} tried to use learning-based approaches to imitate the behavior of the so-called strong branching method, a node-efficient but time-consuming branching variable selection method in the B\&B search tree. In a very recent work by \cite{gasse2019exact}, a GCN model is trained to imitate the strong branching rule. Our model is different from theirs in terms of both the graph and network structure as well as the application scenario of the prediction results. \cite{tang2019reinforcement} designed a deep reinforcement learning framework for intelligent selection of cutting planes. \cite{he2014learning} used imitation learning to train a node selection and a node pruning policy to speed up the tree search in the B\&B process. \cite{khalil2017learning} used binary classification to predict whether a primal heuristic will succeed at a given node and then decide whether to run a heuristic at that node. \cite{kruber2017learning} proposed a supervised learning method to decide whether a Danzig-Wolfe reformulation should be applied and which decomposition to choose among all possibles. Interested readers can refer to \cite{bengio2018machine} for a comprehensive survey on the use of machine learning methods in CO.

The proposed MIP solving framework is different from previous work in two aspects:

\vspace{1mm}
 \noindent\textbf{Generalization}: Previous solution generation method for CO usually focus on problems with certain solution structures. 
For example, applications of Pointer Networks \cite{vinyals2015pointer,kool2018attention} are only suited for sequence-based solution encoding, and reinforcement learning \cite{dai2017learning,li2018combinatorial} type decision making is based on the assumption that a feasible solution can be obtained by sequential decisions. In contrast, the proposed framework does not limit to problems of certain types but applies to most CO problems that can be modeled as MIPs. This greatly enlarges the applicable area of the proposed framework.

\vspace{1mm}
\noindent\textbf{Representation}: Previous applications of ML techniques to enhance MIP solving performance mainly use hand-crafted features, and make predictions on each variable independently. Notice that the solution value of a variable is strongly correlated to the objective function and the constraints it participates in, we build a tripartite graph representation for MIP, based on which graph embedding technique is applied to extract correlations among variables, constraints and the objective function without human intervention.

\section{The Solution Framework}
Consider an MIP problem instance $I$ of the general form:
\begin{align}
\min~~~ &\boldsymbol{c}^T\boldsymbol{x}\\
\text{s.t.}~~~  &\boldsymbol{Ax} \le \boldsymbol{b},\\
& x_j \in \{0,1\}, ~\forall j\in \mathcal{B},\\
& x_j \in \mathbb{Z}, ~\forall j\in \mathcal{Q},~~~x_j \ge 0, ~\forall j\in \mathcal{P},
\end{align}
where the index set of decision variables $\mathcal{U}:=\{1,\ldots,n\}$ is partitioned into $(\mathcal{B}, \mathcal{Q}, \mathcal{P})$, and $\mathcal{B}, \mathcal{Q}, \mathcal{P}$ are the index set of binary, general integer and continuous variables, respectively. 
The main task here is to predict the probability that a binary variable $x_j,~ j\in \mathcal{B}$ to take value 1 (or zero) in the optimal solution. Next, we describe in detail the tripartite graph representation of MIP, the GCN model structure, and how solution prediction results are incorporated to accelerate MIP solving.

\subsection{Graph Representation for MIP}
\begin{figure}[htbp!]
\centering
\includegraphics[width=0.46\textwidth]{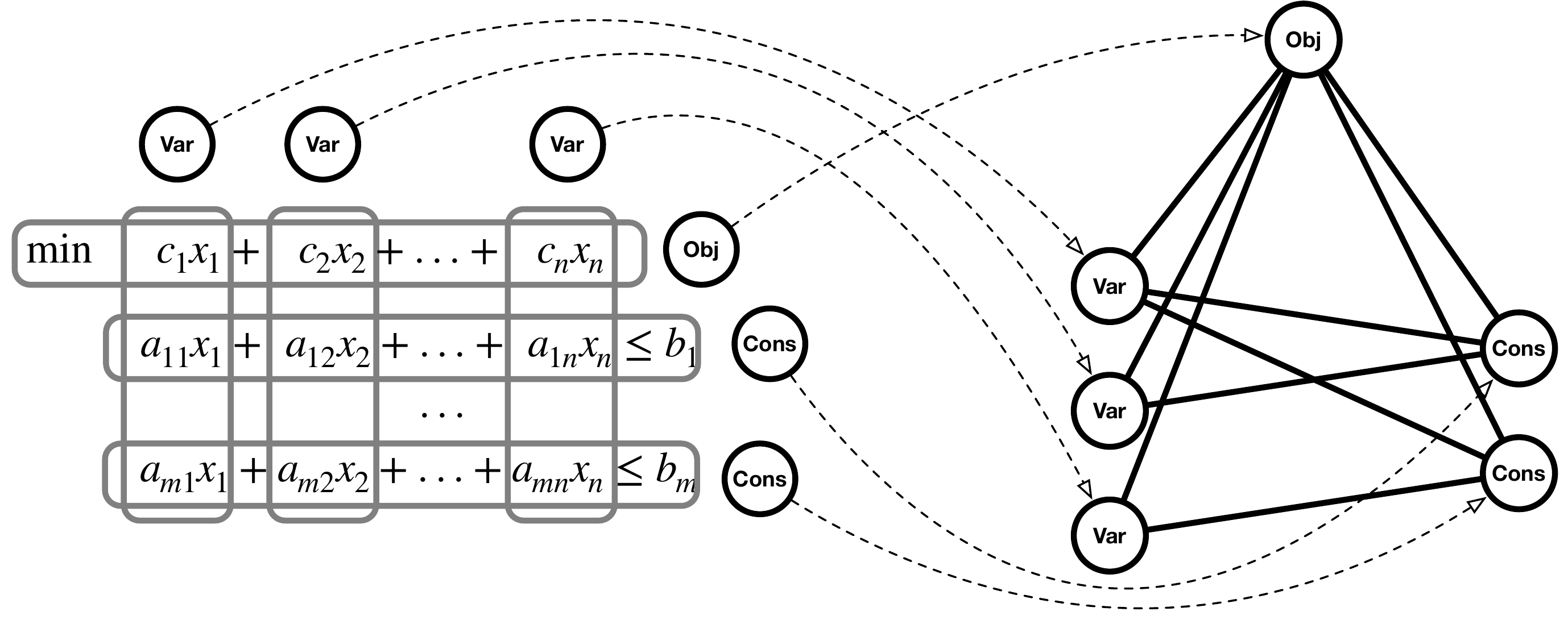}
\caption{Transforming an MIP instance to a tripartiete graph \label{trigraph}}
\end{figure}

Our main idea is to use a tripartite graph $\mathcal{G} = \{\mathcal{V}, \mathcal{E}\}$ to represent an input MIP instance $I$. In particular, objective function coefficients $\boldsymbol{c}$, constraint right-hand-side (RHS) coefficients $\boldsymbol{b}$ and coefficient matrix $\boldsymbol{A}$ information is extracted from $I$ to build the graph. Vertices and edges in the graph are detailed as follows and graphically illustrated in Fig \ref{trigraph}.

\noindent\textbf{Vertices:}\\
\indent 1) the set of decision variable vertices $\mathcal{V}_V$, each of which corresponds to a binary variable in $I$.\\
\indent 2) the set of constraint vertices $\mathcal{V}_C$, each of which corresponds to a constraint in $I$.\\
\indent 3) an objective function vertex $o$. \\

\vspace{-1mm}
\noindent\textbf{Edges:}\\
\indent 1) $v$-$c$ edge: there exists an edge between  $v\in\mathcal{V}_V$ and $c\in\mathcal{V}_C$ if the corresponding variable of $v$ has a non-zero coefficient in the corresponding constraint of $c$ in the MIP formulation.\\
\indent 2) $v$-$o$ edge: for each $v\in\mathcal{V}_V$, there exists an edge between $v$ and $o$.\\
\indent 3) $c$-$o$ edge: for each $c\in\mathcal{V}_C$, there exists an edge between $c$ and $o$.

\begin{remark}
The presented trigraph representation not only captures connections among the variables, constraints and objective functions but maintains the detailed coefficients numerics in its structure as well. In particular, non-zero entries in coefficient matrix $\boldsymbol{A}$ are included as features of $v$-$c$ edges, entries in objective coefficients $\boldsymbol{c}$ as features of $v$-$o$ edges, and entries in $\boldsymbol{b}$ as features of $c$-$o$ edges. Note that the constraint RHS coefficients $\boldsymbol{b}$ are correlated to the objective function by viewing LP relaxation of $I$ from a dual perspective.
\end{remark}

\subsection{Solution Prediction for MIP}
We describe in Algorithm \ref{alg_embedding} the overall forward propagation prediction procedure based on the trigraph. The procedure consists of three stages: 1) a fully-connected ``EMBEDDING'' layer with 64 dimension output for each node so that the node representations are of the same dimension (lines 1-3 in the algorithm). 2) a graph attention network to transform node information among connected nodes (lines 4-12). 3) two fully-connected layers between variable nodes and the output layer (line 13). The sigmoid activation function is used for output so that the output value can be regarded as the probability that the corresponding binary variable takes value 1 in the MIP solution. The overall GCN is trained by minimizing the binary cross-entropy loss.

\begin{algorithm}
\footnotesize
\renewcommand{\algorithmicrequire}{\textbf{Input:}}
\renewcommand{\algorithmicensure}{\textbf{Output:}}
\caption{Graph Convolutional Network (forward propagation)}\label{alg_embedding}
\begin{algorithmic}[1]

\Require Graph $\mathcal{G} = \{\mathcal{V}, \mathcal{E}\}$; Input features $\{\boldsymbol{\rm x}_j, \forall j\in \mathcal{V}\}$; Number of transition iterations $T$; Weight matrices $\boldsymbol{\rm{W}}^t, \forall t\in\{1,\ldots,T\}$ for graph embedding; Output layer weight matrix $\boldsymbol{\rm{W}}^\text{out}$; Non-linearity $\sigma$ (the relu function); Non-linearity $\sigma_s$ (the sigmoid function); Neighborhood function $\mathcal{N}$; Attention coefficients $\boldsymbol{\alpha}$.

\Ensure Predicted value of binary variables: $z_v, \forall v\in\mathcal{V}_V$.
\State $\boldsymbol{h}_v^{0}\gets \text{EMBEDDING}(\boldsymbol{\rm x}_v), \forall v\in\mathcal{V}_V$
\State $\boldsymbol{h}_c^{0}\gets \text{EMBEDDING}(\boldsymbol{\rm x}_c), \forall c\in\mathcal{V}_C$
\State $\boldsymbol{h}_o^{0}\gets \text{EMBEDDING}(\boldsymbol{\rm x}_o)$
\For {$t = 1,\ldots, T$}
	\State $\boldsymbol{h}_o^{t} \gets \sigma\Big(\boldsymbol{\rm{W}}_{Vo}^t\cdot\text{CONCAT}\big(\boldsymbol{h}_o^{t-1},\ \sum\limits_{v\in\mathcal{V}_V\cap \mathcal{N}(o)}\alpha_{vo}\boldsymbol{h}_v^{t-1}\big)\Big)$
	\hspace{-0.2cm}\For {$c$ in $\mathcal{C}}:$
		\State \hspace{-0.1cm}$\boldsymbol{h}_o^{t}\gets \sigma\Big(\boldsymbol{\rm{W}}^t_{o{C}}\cdot\text{CONCAT} \left(\boldsymbol{h}_o^{t}, \boldsymbol{h}_c^{t-1}\right)\Big)$
		\State \hspace{-0.1cm}$\boldsymbol{h}_c^{t} \gets \sigma\Big(\boldsymbol{\rm{W}}^t_{{V}{c}}\cdot\text{CONCAT}\big(\boldsymbol{h}_o^{t}, \sum\limits_{v\in\mathcal{V}_V\cap \mathcal{N}(c)}\alpha_{vc}\boldsymbol{h}_v^{t-1}\big)\Big)$
	\EndFor
		\State $\boldsymbol{h}_o^{t} \gets \sigma\Big(\boldsymbol{\rm{W}}^t_{{C}o}\cdot\text{CONCAT}\big(\boldsymbol{h}_o^{t},\ \sum\limits_{c\in\mathcal{V}_C					\cap\mathcal{N}(o)}\alpha_{co}\boldsymbol{h}_c^{t}\big)\Big)$
	\For {$v$ in $\mathcal{V}}:$
		\State \hspace{-0.1cm}$\boldsymbol{h}_o^{t}\gets \sigma\Big(\boldsymbol{\rm{W}}^t_{o{V}}\cdot\text{CONCAT}\left(\boldsymbol{h}_o^{t}, \boldsymbol{h}_v^{t-1}\right)\Big)$
		\State \hspace{-0.1cm}$\boldsymbol{h}_v^{t} \gets \sigma\Big(\boldsymbol{\rm{W}}^t_{{C}{v}}\cdot\text{CONCAT}\big(\boldsymbol{h}_o^{t},\ \sum\limits_{c\in\mathcal{V}_C\cap \mathcal{N}(v)}\alpha_{cv}\boldsymbol{h}_c^{t}\big)\Big)$
	\EndFor	
\EndFor
\State $z_v \gets \sigma_s\left(\boldsymbol{\rm{W}}^{\text{out}}\cdot \text{CONCAT}\left(\boldsymbol{h}_v^{0}, \boldsymbol{h}_v^{T}\right)\right), \forall v\in\mathcal{V}_V$
\end{algorithmic}
\end{algorithm}

\begin{figure}[htbp!]
\begin{center}
\includegraphics[width=0.45\textwidth]{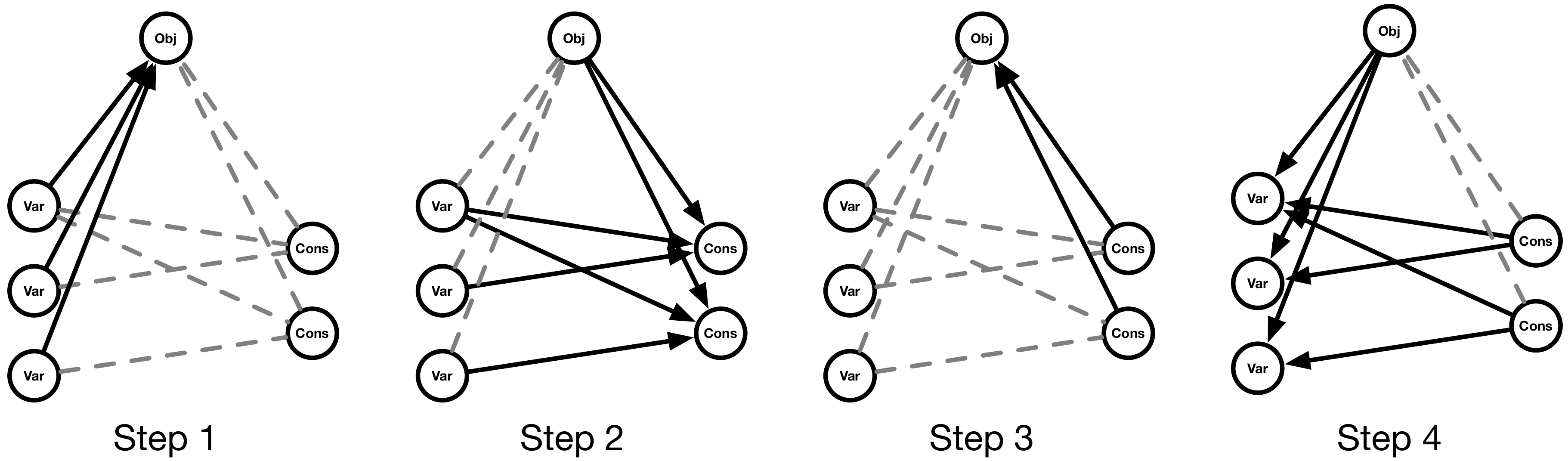}
\caption{Information transition flow in the trigraph convolutional layer\label{infoFlow}}
\end{center}
\tiny 
\noindent The information transitions run consecutively as follows: Step 1, transform variable nodes information to the objective node; Step 2, transform the objective and variable nodes information to constraint nodes; Step 3, transform constraint nodes information to the objective node; Step.4, transform the objective node and constraint nodes information to variable nodes; 
\end{figure}

Nodes' representations in the tripartite graph are updated via a 4-step procedure. In the first step (line 5 in Algorithm \ref{alg_embedding}), the objective node $o$ aggregates the representations of all variable nodes $\{\boldsymbol{h}_v, v\in\mathcal{V}_V\}$ to update its representation  $\boldsymbol{h}_o$. The ``CONCAT'' operation represents the CONCATENATE function that joins two arrays. In the second step (lines 6-8), $\{\boldsymbol{h}_v, v\in\mathcal{V}_V\}$ and $\boldsymbol{h}_o$ are used to update the representations of their neighboring constraint node $c\in\mathcal{V}_C$. In the third step (line 9), representations of constraints $\{\boldsymbol{h}_c, c\in\mathcal{V}_C\}$ are aggregated to update $\boldsymbol{h}_o$, while in the fourth step (lines 10-12), $\{\boldsymbol{h}_c, c\in\mathcal{V}_C\}$ and $\boldsymbol{h}_o$ are combined to update $\{\boldsymbol{h}_v, v\in\mathcal{V}_V\}$.
 See Fig.~\ref{infoFlow} for an illustration of information transition flow in the trigraph. After $T$ transitions, two fully-connected layers coupled with a sigmoid activation function $\sigma_s$ is used for solution value prediction of each $v\in\mathcal{V}_V$ (line 13).

The intuition behind Algorithm \ref{alg_embedding} is that at each iteration, a variable node incrementally gathers more aggregation information from its neighboring nodes, which correspond to the related constraints and variables in the MIP formulation. It is worth mentioning that these transitions only extract connection relations among the nodes, ignoring the detailed coefficients numerics $\boldsymbol{A,b}$ and $\boldsymbol{c}$.  To enhance the representation ability of our model, we include an attention mechanism to import information from the coefficients' values.

\vspace{2mm}
\noindent\textbf{Attention Mechanism:} A distinct feature in the tripartite graph structure is the heterogeneities in nodes and arcs. Rather than using a shared linear transformation (i.e., weight matrix) for all nodes \cite{velivckovic2017graph}, we consider different transformations in each step of graph embedding updates, reflecting the importance of feature of one type of nodes on the other. In particular, given node $i$ of type ${T}_i$ and node $j$ of type ${T}_j$, the attention coefficient which indicates the importance of node $i\in\mathcal{V}$ from its neighbor $j\in\mathcal{N}(i)$ is computed as:
\begin{align}
\alpha_{ij} = \sigma_s\left(\boldsymbol{\rm W}_{{T}_i,{T}_j}^\text{att}\cdot\text{CONCAT}\left(\boldsymbol{h}_{i},  \boldsymbol{h}_{e_{ij}},  \boldsymbol{h}_{j}\right)\right),
\end{align}
where $\boldsymbol{h}_{i},  \boldsymbol{h}_{j}, \boldsymbol{h}_{e_{ij}}$ are embeddings of node $i,j\in\mathcal{V}$ and edge $(i,j)\in\mathcal{E}$ respectively, $\sigma_s$ is the sigmoid activation function, and $\boldsymbol{\rm W}_{{T}_i,{T}_j}^\text{att}$ is the attention weight matrix between type ${T}_i$ and ${T}_j$ nodes. For each $i\in\mathcal{V}$, the attention coefficient is normalized cross over all neighbor nodes $j\in\mathcal{N}(i)$ using a softmax function.
With this mechanism, coefficients information in $\boldsymbol{A,b}$ and $\boldsymbol{c}$ (all of which contained in the feature vector of the edges) are incorporated to reflect edge connection importance in the graph.

\subsection{Prediction-Based MIP Solving}
\label{chap_MIPsolving}
Next, we introduce how the solution value prediction results are utilized to improve MIP solving performance. One approach is to add a local branching \cite{fischetti2003local} type (invalid) global cut to the MIP model to reduce the search space of feasible solutions. This method aims to identify decision variables that are predictable and stable, and restrict the B\&B tree search on unpredictable variables to accelerate primal solution-finding. An alternative approach is to perform an actual branching at the root node that maintains global optimality. These two methods are detailed as follows.

\vspace{-1mm}
\subsubsection{a) Approximate approach.}
Let $\hat x_j$ denote the predicted solution value of binary variable $x_j, j\in \mathcal{B}$, and let $\mathcal{S}\subseteq \mathcal{B}$ denote a subset of indices of binary variables. A local branching initial cut to the model is defined as:
\begin{equation}
\label{eq_phi}
\Delta(\boldsymbol{x}, \hat{\boldsymbol{x}}, \mathcal{S}) = \sum_{j\in \mathcal{S}:\hat x_j^k = 0}x_j + \sum_{j\in \mathcal{S}:\hat x_j^k = 1}(1-x_j) \le \phi,
\end{equation}
where $\phi$ is a problem parameter that controls the maximum distance from a new solution $\boldsymbol{x}$ to the predicted solution $\boldsymbol{\hat x}$. Adding cuts with respect to subset $\mathcal{S}$ rather than $\mathcal{B}$ is due to the unpredictable nature of unstable variables in MIP solutions. Therefore, only those variables with high probability to take value 0 or 1 are included in $\mathcal{S}$. For the extreme case that $\phi$ equals 0, the initial cut is equivalent to fixing variables with indices in $\mathcal{S}$ at their predicted values. It is worth mentioning that the inclusion of global constraint (\ref{eq_phi}) shrinks the feasible solution region of the original model, trading optimality for speed as an approximate approach.

\vspace{-1mm}
\subsubsection{b) Exact approach.} The proposed local branching type cut can also be incorporated in an exact solver by branching on the root node. To do this, we create two child nodes from the root as follows:
$$\text{Left: }~\Delta(\boldsymbol{x}, \hat{\boldsymbol{x}}, \mathcal{S}) \le \phi,  ~~~~~\text{Right: } ~ \Delta(\boldsymbol{x}, \hat{\boldsymbol{x}}, \mathcal{S}) \ge \phi + 1,$$
which preserves all feasible solution regions in the tree search. After the root node branching, we switch back to the solver's default setting and perform an exact B\&B tree search process.


\section{Data Collection}
\textbf{Features:} An ideal feature collection procedure should capture sufficient information to describe the solution process, and being of low computational complexity as well. A good trade-off between these two concerns is to collect features at the root node of the B\&B tree, where the problem has been presolved to eliminate redundant variables and constraints and the LP relaxation is solved.  In particular, we collect for each instance 3 types of features: variable features, constraint features, and edge features. Features descriptions are summarized in Table 1 in Appendix A.

As presented in the feature table, features of variables and constraints can be divided into three categories: basic features, LP features, and structure features. The structure features (most of which can be found in \cite{alvarez2017machine,khalil2016learning}) are usually hand-crafted statistics to reflect correlations between variables and constraints. It is noticeable that our tripartite graph neural network model can naturally capture these correlations without human expertise and can generate more advanced structure information to improve prediction accuracy. This will be verified in the computational evaluations section.\\

\vspace{-2mm}
\noindent\textbf{Labels:} To make predictions on solution values of binary variables, an intuitive labeling scheme for the variables is to label them with the optimal solution values. Note, however, obtaining optimal solutions for medium or large scale MIP instances can be very time-consuming or even an impossible task. This implies that labeling with optimal solutions can only be applied to solvable MIP instances, which limits the applications of the proposed framework. 

In the situation when optimal solutions are difficult to obtain, we propose to identify \emph{stable} and \emph{unstable} variables in solutions. This is motivated by the observation that solution values of the majority of binary decision variables remain unchanged across a series of different feasible solutions. To be specific, given a set of $K$ solutions $\{\boldsymbol{\bar x}^1, \ldots, \boldsymbol{\bar x}^K\}$ to an MIP instance $I$, a binary variable $x_j$ is defined as unstable if there exists some $k_1, k_2\in\{1, \ldots, K\}$ such that $x_j^{k_1}\neq x_j^{k_2}$, and as stable otherwise. Although obtaining optimal solutions might be a difficult task, the stable variables can be viewed as an easy-to-predict part in the (near) optimal solutions. To generate a sequence of solutions to an instance, we use the proximity search method \cite{fischetti2014proximity}. Starting from some initial solution $\boldsymbol{\bar x}^k$ with objective value $\boldsymbol{c}^T\boldsymbol{\bar x}^k$, a neighborhood solution with the objective value improvement being at least $\delta$ can be generated by solving the following optimization:
\begin{align}
\min~~ &\sum_{j\in \mathcal{B}:\bar x_j^k = 0}x_j + \sum_{j\in \mathcal{B}:\bar x_j^k = 1}(1-x_j)\\
\text{s.t.}~~  & \boldsymbol{c}^T\boldsymbol{x}\le \boldsymbol{c}^T\boldsymbol{\bar x}^k - \delta,\\
&\boldsymbol{Ax}\le \boldsymbol{b},\\
& x_j \in \{0,1\}, \forall j\in \mathcal{B},\\
&x_j \in \mathbb{Z}, \forall j\in \mathcal{G}; ~~ x_j \ge 0, \forall j\in \mathcal{C},
\end{align}
where the objective function represents the distance between $\boldsymbol{\bar x}^k$ and a new solution $\boldsymbol{x}$. Note that the above optimization is computationally tractable since solving process can terminate as soon as a feasible solution is found. By iteratively applying this method, we obtain a series of improving feasible solutions to the original problem. 
Stable binary variables are labeled with their solution values while the unstable variables are marked as unstable and discarded from the training set. A limitation of this labeling method is the inability of handling the case when the initial feasible solution $\boldsymbol{\bar x}^0$ is hard to obtain.

\begin{remark}
The logic behind the stable variable labeling scheme is to explore local optimality patterns when global optimality is not accessible. In each iteration of proximity search, a neighboring better solution is found, with a few flips on solution values of the binary variables. Performing this local search step for many rounds can identify local minimum patterns which reflect domain knowledge of the CO problem. Take the Traveling Salesman Problem (TSP) as an example. Let $z_{jl}$ define whether node $l$ is visited immediately after node $j$. If $j$ and $l$ are geometrically far away from each other, $z_{jl}$ is likely to be zero in all the solutions generated by proximity search and being recorded by our labeling scheme, reflecting the underlying local optimality knowledge for TSP. 
\end{remark}

\section{Experimental Evaluations}
\textbf{Setup.} To evaluate the proposed framework, we modify the state-of-the-art open-source MIP solver SCIP (version 6.0.1) for data collection and solution quality comparison. The GCN model is built using the Tensorflow API. All experiments were conducted on a cluster of three 4-core machines with Intel 2.2 GHz processors and 16 GB RAM.\\

\vspace{-2.5mm}
\noindent\textbf{Instances.} To test the effectiveness and generality of the prediction-based solution framework, we generate MIP instances of 8 distinct types: \emph{Fixed Charge Network Flow} (FCNF), \emph{Capacitated Facility Location} (CFL), \emph{Generalized Assignment} (GA), \emph{Maximal Independent Set} (MIS), \emph{Multidimensional Knapsack} (MK), \emph{Set Covering} (SC), \emph{Traveling Salesman Problem} (TSP) and \emph{Vehicle Routing Problem} (VRP). These problems are the most commonly visited NP-hard combinatorial optimizations in OR and are quite general because they differ significantly in MIP structures and solution structures. For each problem type, 200 MIP instances of similar scales are generated. The number of instances used for training, validation, and testing is 140, 20, 40 respectively. Parameter calibrations are performed on the validation set, while prediction accuracy and solution quality comparisons are evaluated on the test instances. Detailed MIP formulation and instance parameters of each type are included in Appendix B. \\

\vspace{-2.5mm}
\noindent\textbf{Data collection.} In terms of feature collection, we implemented a feature extraction plugin embedded in the branching procedure of SCIP. In particular, variable features, constraint features, and edge features are collected right before the first branching decision is made at the root node, where the presolving process, root LP relaxation, and root cutting plane have completed. No further exploration of the B\&B search tree is needed and thus the feature collection process terminates at the root node. Construction of the tripartite graph is also completed at the root node where SCIP is working with a transformed MIP model such that redundant variables and constraints have been removed. In terms of label collection, since optimal solutions to all problem types can not be obtained within a 10000 seconds time limit,  we applied the proximity search method to label stable binary variables. The initial solution $\boldsymbol{\bar x}^0$ for proximity search is obtained under SCIP's default setting with a 300 seconds execution time limit. If SCIP fails to obtain an initial feasible solution within the time limit, the time limit doubles until a feasible initial solution can be found. Parameter $\delta$ for the proximity search is set as $0.01\cdot (\boldsymbol{c}^T\boldsymbol{\bar x}^0 - \text{LB})$ where \text{LB} is the lower bound objective value given by the solver. Each proximity search iteration terminates as soon as a feasible solution is found. This process generally converges within 20 to 40 iterations.\\

\vspace{-2.5mm}
\noindent\textbf{Parameter calibration.} The performance of the proposed framework benefits from a proper selection of hyper-parameters $\phi$ and $\mathcal{S}$. Let $z_j$ denote the prediction probability that binary variable $x_j, j\in\mathcal{B}$ takes value 1 in the MIP solution. We sort $x_j$ in non-decreasing order of $\min(z_j, 1-z_j)$ and choose the first $\eta\cdot|\mathcal{B}|$ variables as $\mathcal{S}$. The strategy of tuning $\phi\in\{0, 5, 10, 15, 20\}$ and $\eta\in\{0.8, 0.9, 0.95, 0.99, 1\}$ is grid search, where the combination of $\phi$ and $\eta$ that results in best solution quality on the validation instances is selected. Table \ref{table_parameter} summarizes $\phi$ and $\eta$ selections for each problem type.

\vspace{-1mm}
\begin{table}[htbp!]
\centering
\footnotesize
\caption{Hyper-parameter selections for each problem type \label{table_parameter}}
\vspace{-2mm}
\begin{threeparttable}
\begin{tabular}{@{}lcccccccc@{}}
\toprule
& FCNF & CFL & GA & MIS & MK & SC & TSP & VRP \\ \midrule
$\phi$  & 0   & 0  & 5   & 10   & 10   & 0    &0   &  5    \\
$\eta$  & 0.80 & 0.95 & 0.99 & 0.90 & 0.80 & 0.90 & 0.90 & 0.95\\
\bottomrule
\end{tabular}
\end{threeparttable}
\end{table}

\subsection{Results of Solution Prediction}
We demonstrate the effectiveness of the proposed GCN model on prediction accuracy against the XGBoost (XGB) classifier \cite{chen2016xgboost}. For XGB, only variable features are used for prediction since it can not process the trigraph information. Noting that solution values of binary variables are usually highly imbalanced,  we use the \emph{average precision} (AP) metric \cite{zhu2004recall} to evaluate the performance of the classifiers. In particular, the AP value is defined as:
\begin{equation}
\text{AP}=\sum_{k=1}^n{P(k)\Delta r(k)},
\end{equation}
where $k$ is the rank in the sequence of predicted variables, $P(k)$ is the precision at cut-off $k$ in the list, and $\Delta r(k)$ is the difference in recall from $k-1$ to $k$.

\begin{table}[htbp!]
 \caption{Comparisons on the average precision metric\label{table_pred}}
 \vspace{-2mm}
\centering
  \footnotesize
  \begin{tabular}{@{}ccccccccccccc@{}}
  \toprule
          & \multicolumn{2}{c}{Basic} &  \multicolumn{2}{c}{Basic\&structure}  &    \multicolumn{2}{c}{All}  \\ \cmidrule(lr){2-3} \cmidrule(lr){4-5} \cmidrule(lr){6-7}          
	Instances&	XGB	&	GCN	&	XGB	&	GCN&	XGB	&	GCN	\\	\midrule
FCNF	&	0.099	&	\textbf{0.261}	&	0.275	&	\textbf{0.317}	&	0.787	&	\textbf{0.788}\\	
CFL	&	0.449	&	\textbf{0.590}	&	0.567	&	\textbf{0.629}	&	0.846	&	\textbf{0.850}\\	
GA	&	0.499	&	\textbf{0.744}	&	0.750	&	\textbf{0.797}	&	0.936	&	\textbf{0.937}\\	
MIS	&	0.282	&	\textbf{0.355}	&	0.289	&	\textbf{0.337}	&	0.297	&	\textbf{0.325}\\	
MK	&	0.524	&	\textbf{0.840}	&	0.808	&	\textbf{0.843}	&	0.924	&	\textbf{0.927}\\
SC	&	0.747	&	\textbf{0.748}	&	0.748	&	\textbf{0.753}	&	\textbf{0.959}	&	\textbf{0.959}\\	
TSP	&	0.327	&	\textbf{0.358}	&	0.349	&	\textbf{0.353}	&	0.401	&	\textbf{0.413}\\
VRP	&	0.391	&	\textbf{0.403}	&	0.420	&	\textbf{0.424}	&	0.437	&	\textbf{0.459}\\	
\midrule
Average   & 0.415  &  \textbf{0.537}  & 0.526  &  \textbf{0.556}   & 0.698  &  \textbf{0.707}\\
  \bottomrule   
  \end{tabular}
\end{table}

Table \ref{table_pred} describes the AP value comparison results for the two classifiers under three settings: using only basic features, using basic\&structure features, and using all features respectively. It is observed that the proposed GCN model outperforms the baseline classifier in all settings. The performance advantage is particularly significant in the basic feature columns (0.537 by GCN against 0.415 by XGB), where only raw coefficient numerics in MIP are used for prediction. The other notable statistic in the table is that the GCN model with only basic features is on average superior to the XGB classifier with basic\&structure features, indicating that the proposed embedding framework can extract more information compared to hand-crafted structure features used in the literature \cite{alvarez2017machine,khalil2016learning}. For comparisons in the all features column, the advantage of GCN is less significant due to the reason that high-level MIP structure information is also captured in its LP relaxations.

\begin{figure}[htbp!]
\centering
\includegraphics[width=0.35\textwidth]{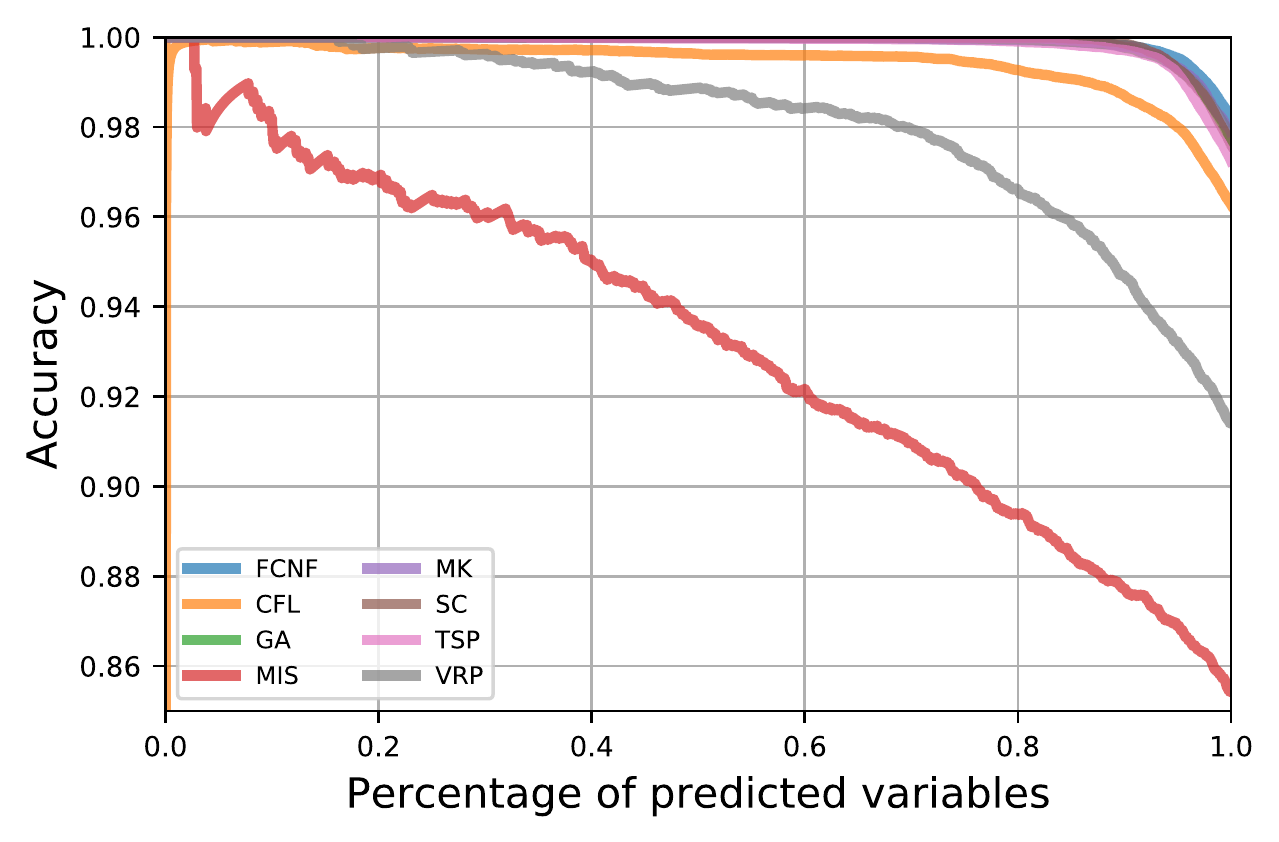}
\caption{Accuracy under different prediction percentage\label{Fig_acc_frac}}
\end{figure}

\vspace{-2mm}
To help illustrate the predictability in solution values for problems of different types,  we present in Fig.\ref{Fig_acc_frac} the detailed accuracy curve for each of the 8 problem types using the GCN model with all features. The figure depicts the prediction accuracy if we only predict a certain percentage of the most predictable binary variables\footnote{The predictability of a binary variable $x_j$ is measured by $\max(z_j, 1-z_j)$.}.  It can be observed from the figure that solution values of most considered MIP problems (such as FCNF, GA, MK, SC,TSP) are fairly predictable, with an almost 100\% accuracy if we make solution value prediction on the top 0-80\% most predictable variables. Among the tested problem types, solution values to MIS and VRP problem instances are fairly unpredictable, which is consistent with the hyper-parameter calibration outcomes that $\phi$ takes value greater than 0.


\subsection{Comparisons of solution quality}
\subsubsection{a) Approximate approach:}
To evaluate the value of incorporating the (invalid) global cut in MIP solving,  we compare the performance of prediction-based approximate approach against that of the solver's aggressive heuristics setting\noindent\footnote{As far as we know, there are hardly any stable approximate solvers for MIP and ``set heuristics emphasis aggressive'' is the most relevant setting we find to accelerate primal solution-finding in the SCIP's documentation. Therefore we use this setting as a benchmark for comparison.}. To be specific, we modified SCIP's setting to ``set heuristics emphasis aggressive'' to make SCIP focus on solution finding rather than proving optimality. For each problem type, 10 MIP instances are randomly selected from the 40 testing instances for solution quality comparisons. 

Notice that the proposed approximate approach does not guarantee global optimality (i.e., does not provide a valid lower bound), we use the primal gap metric \cite{khalil2017learning} to capture solver's performance on primal solution finding. In particular, the \emph{primal gap} metric $\gamma(\tilde{\boldsymbol{x}})$ reports the relative gap in the objective value of a feasible solution $\tilde{\boldsymbol{x}}$ to that of the optimal (or best-known) solution $\tilde{\boldsymbol{x}}^*$:
\begin{equation}
\gamma(\tilde{\boldsymbol{x}}) = \frac{|\boldsymbol{c}^T\tilde{\boldsymbol{x}} - \boldsymbol{c}^T\tilde{\boldsymbol{x}}^* |}{\max\{|\boldsymbol{c}^T\tilde{\boldsymbol{x}}|, | \boldsymbol{c}^T\tilde{\boldsymbol{x}}^* |\} + \epsilon}\times 100\%,
\end{equation}
where $\epsilon = 10^{-10}$ is a small constant to avoid numerical error when $\max\{|\boldsymbol{c}^T\tilde{\boldsymbol{x}}|, | \boldsymbol{c}^T\tilde{\boldsymbol{x}}^* |\}=0$. Since all problem types are not solvable within a 10000 seconds time limit (under the SCIP's default setting without the global cut), $\tilde{\boldsymbol{x}}^*$ is selected as the best-known solution found by all tested methods.

\begin{table}[htbp!]
\footnotesize
\caption{Primal gap comparisons for the approximate approach (\%)\label{table_sol}}
\vspace{-2mm}
\centering
\begin{threeparttable}
\begin{tabular}{crrrrrr}
  \toprule
   & AGG1\tnote{*}  & AGG2 & AGG5 & AGG10  & GCN-A    \\
     \midrule  
FCNF & 1.733 & 1.733 & 1.678 & 0.380 & \textbf{0.000} &\\
CFL & 2.334 & 2.112 & 0.868 & 0.206 & \textbf{0.092} &\\
GA & 0.451 & 0.430 & 0.430 & 0.430 & \textbf{0.000} &\\
MIS & 10.292 & 6.125 & 5.083 & 5.083 & \textbf{1.042} &\\
MK & 0.003 & 0.003 & \textbf{0.000} & \textbf{0.000} & 0.003 &\\
SC & 1.509 & 0.529 & 0.383 & \textbf{0.000} & 0.164 &\\
TSP & 10.387 & 6.286 & 2.752 & 1.981 & \textbf{0.332} &\\
VRP & 3.096 & 3.096 & 1.177 & 1.131 & \textbf{0.896} &\\
\midrule
Average &	 3.726 & 2.539 & 1.546 &	1.151 	& \textbf{0.316} 	\\		
  \bottomrule  
\end{tabular}
\begin{tablenotes}
\tiny
\item[*] AGG1 represents SCIP's aggressive heuristics setting with $1\times 1000$ seconds execution time limit. Similarly, AGG2, AGG5 and AGG10 correspond to aggressive heuristics setting with $2\times 1000$, $5\times 1000$ and $10\times 1000$ seconds time limit respectively.
      \end{tablenotes}
\end{threeparttable}
\end{table}

Table \ref{table_sol} is a collection of solution quality results by the proposed method with a 1000 seconds execution time limit. To demonstrate the significance of performance improvement, we allow the solver to run for 2-10 times the execution time (i.e., 2000-10000 seconds) in aggressive heuristics setting. It is revealed from the table that the proposed approximate method (the GCN-A column) gains remarkable advantages to the SCIP's aggressive heuristics setting under the same time limit (the AGG1 column) on all testing problems. Compared to the setting with 10000 seconds (the AGG10 column), the proposed framework still maintains an average better performance, indicating a 10 times acceleration in solution-finding, with the trade of optimality guarantees. 

\subsubsection{b) Exact approach} 
To evaluate the value of performing an actual branching at the root,  we compare the performance of the prediction-based exact approach against that of the solver's default setting with a 1000 seconds time limit. Because the exact approach provides a valid lower bound to the original problem, we use the well-known \emph{optimality gap} metric $\zeta(\tilde{\boldsymbol{x}}, \text{LB})$ to measure the overall MIP solving performance:
\begin{equation}
\zeta(\tilde{\boldsymbol{x}}, \text{LB}) = \frac{|\boldsymbol{c}^T\tilde{\boldsymbol{x}} - \text{LB}|}{|\boldsymbol{c}^T\tilde{\boldsymbol{x}}|+ \epsilon}\times 100\%,
\end{equation}
where $\tilde{\boldsymbol{x}}$ and $\text{LB}$ denote respectively the primal solution and best lower bound obtained by a specific method.

\begin{table}[htbp!]
\centering
\footnotesize
\caption{Optimality gap comparisons for the exact approach (\%)\label{table_exact}}
\vspace{-2mm}
\begin{threeparttable}
\begin{tabular}{@{}L{10mm}R{5mm}R{5mm}R{4mm}R{5mm}R{4mm}R{5mm}R{5mm}r@{}}
\toprule
& FCNF & CFL & GA & MIS & MK & SC & TSP & VRP \\ \midrule
DEF  & 7.06   & 3.96   & 5.30   & 62.92   & 2.01  & 12.09   & \textbf{12.13}  & 68.69    \\
GCN-E  & \textbf{4.78}   & \textbf{3.64}   & \textbf{4.31}   & \textbf{61.75}   & \textbf{2.00}  & \textbf{11.19}   & 12.84  & \textbf{68.13} \\
\bottomrule
\end{tabular}
\end{threeparttable}
\end{table}

We conclude the experimental results in Table \ref{table_exact}. The DEF row corresponds to SCIP's default setting and GCN-E corresponds to the exact approach using the new root branching rule. It is revealed from the table that GCN-E outperforms DEF in terms of the optimality gap within the time limit. This provides empirical evidence that the proposed method is potentially useful for accelerating MIP to global optimality. 

\begin{figure}[htbp!]
\centering
\includegraphics[width=0.47\textwidth]{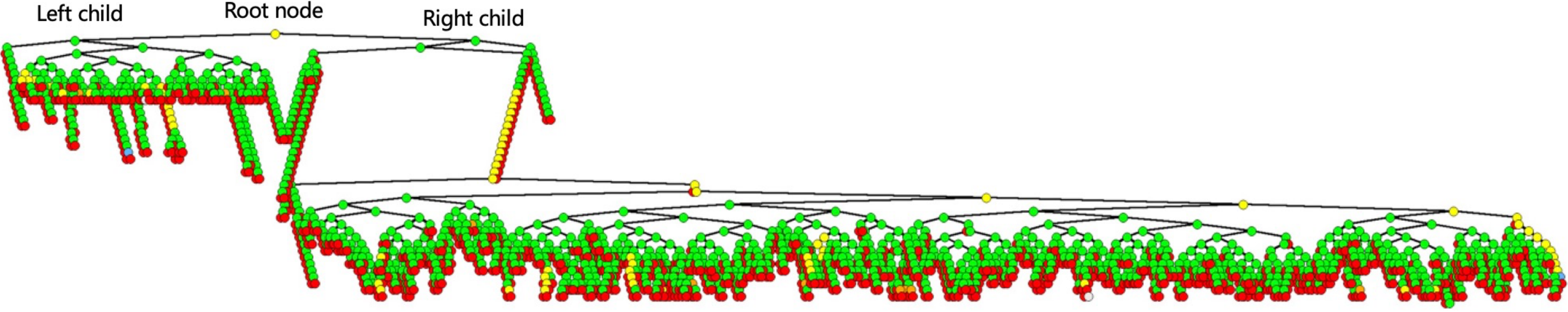}
\caption{Visualization of the B\&B tree after performing a root branching based on GCN prediction. \label{fig_BBtree}}
\end{figure}

To help understand how the new branching rule accelerates the MIP solving process, we plot the B\&B tree of a small-scale ``CFL'' instance in Fig.~\ref{fig_BBtree}. It is observed from the tree search process that the solver finds a good solution quickly on the left tree, and goes to the right tree to get the optimal solution, and finally prove optimality by exploring the remaining nodes on both sides.

\subsection{Generalization to larger instances}
The graph embedding framework endows the model to train and test on MIP instances of different scales. This is important for MIP solving since there is hardly any good strategy to handle large-scale NP-hard MIP problems. To investigate this, we generate 200 small-scale MIP instances for each problem type and train our GCN model on these instances and test its applicability in large-scale ones. Detailed statistics of small and large instances are reported in Appendix B. It is revealed from table \ref{table_general} that the GCN model maintains an acceptable prediction accuracy degradation when the problem scale differs in the training and testing phase. Besides, the prediction result is still useful to improve solver's primal solution finding performance. 

\vspace{-1mm}
\begin{table}[htbp!]
\centering
  \footnotesize
  \caption{Generalization ability of the proposed framework \label{table_general}}
  \vspace{-2mm}
  \begin{threeparttable}
    \begin{tabular}{ccc|ccc}
    \toprule
    & \multicolumn{2}{c}{Average precision} &  \multicolumn{3}{c}{Primal gap (\%)}  \\ \cmidrule(lr){2-3} \cmidrule(lr){4-6}          
       &  GCN   &  GCNG\tnote{*}  &  AGG1   &  GCN-A   &  GCNG-A  \\ 
       \midrule
FCNF  & 0.653  & 0.675  & 1.733  & 0.000  & 2.119  \\ 
CFL  & 0.837  & 0.801  & 2.341  & 0.100  & 0.410  \\ 
GA  & 0.963  & 0.873  & 0.661  & 0.211  & 0.016  \\ 
MIS  & 0.091  & 0.104  & 2.223  & 1.136  & 1.087  \\ 
MK  & 0.789  & 0.786  & 0.000  & 0.000  & 0.000  \\ 
SC  & 0.878  & 0.843  & 1.349  & 0.000  & 0.215  \\ 
TSP  & 0.396  & 0.343  & 10.061  & 0.000  & 4.802  \\ 
VRP  & 0.358  & 0.321  & 4.408  & 1.254  & 1.239  \\ 
\hline
Average  & 0.621  & 0.593  & 2.847  & 0.338  & 1.236 \\ 
  \bottomrule
  \end{tabular}
  \begin{tablenotes}
        \tiny
\item[*] GCNG is the GCN model trained on small-scale MIP instances. GCNG-A is the approximate solving approach based on the GCNG prediction model.
      \end{tablenotes}
  \end{threeparttable}
  \end{table}


\section{Conclusions}
We presented a supervised solution prediction framework to explore the correlations between the MIP formulation structure and its local optimality patterns. The key feature of the model is a tripartite graph representation for MIP, based on which graph embedding is used to extract connection information among variables, constraints and the objective function. Through extensive experimental evaluations on 8 types of distinct MIP problems, we demonstrate the effectiveness and generality of the GCN model in prediction accuracy. Incorporated in a global cut to the MIP model, the prediction results help to accelerate SCIP's solution-finding process by 10 times on similar problems with a sacrifice in proving global optimality. This result is inspiring to practitioners who are facing routinely large-scale MIPs on which the solver's execution time is unacceptably long and tedious, while global optimality is not a major concern.

Limitations of the proposed framework are two-fold. First, this method is better being applied to binary variable intensive MIP problems due to the difficulties in solution value prediction for general integer variables. Second, the prediction performance degrades for problems without local optimality structure where correlations among variables from the global view can not be obtained from the neighborhood information reflected in the MIP's trigraph representation.


\bibliography{myBib}

\begin{thebibliography}{}

\bibitem[\protect\citeauthoryear{Alvarez, Louveaux, and
  Wehenkel}{2017}]{alvarez2017machine}
Alvarez, A.~M.; Louveaux, Q.; and Wehenkel, L.
\newblock 2017.
\newblock A machine learning-based approximation of strong branching.
\newblock {\em INFORMS Journal on Computing} 29(1):185--195.

\bibitem[\protect\citeauthoryear{Alvarez, Wehenkel, and
  Louveaux}{2016}]{marcos2016online}
Alvarez, A.~M.; Wehenkel, L.; and Louveaux, Q.
\newblock 2016.
\newblock Online learning for strong branching approximation in
  branch-and-bound.

\bibitem[\protect\citeauthoryear{Bengio, Lodi, and
  Prouvost}{2018}]{bengio2018machine}
Bengio, Y.; Lodi, A.; and Prouvost, A.
\newblock 2018.
\newblock Machine learning for combinatorial optimization: a methodological
  tour d'horizon.
\newblock {\em arXiv preprint arXiv:1811.06128}.

\bibitem[\protect\citeauthoryear{Chen and Guestrin}{2016}]{chen2016xgboost}
Chen, T., and Guestrin, C.
\newblock 2016.
\newblock Xgboost: A scalable tree boosting system.
\newblock In {\em Proceedings of the 22nd acm sigkdd international conference
  on knowledge discovery and data mining},  785--794.
\newblock ACM.

\bibitem[\protect\citeauthoryear{Chen}{2010}]{chen2010integrated}
Chen, Z.-L.
\newblock 2010.
\newblock Integrated production and outbound distribution scheduling: review
  and extensions.
\newblock {\em Operations research} 58(1):130--148.

\bibitem[\protect\citeauthoryear{Clarke and
  Wright}{1964}]{clarke1964scheduling}
Clarke, G., and Wright, J.~W.
\newblock 1964.
\newblock Scheduling of vehicles from a central depot to a number of delivery
  points.
\newblock {\em Operations research} 12(4):568--581.

\bibitem[\protect\citeauthoryear{Cook}{2011}]{cook2011pursuit}
Cook, W.~J.
\newblock 2011.
\newblock {\em In pursuit of the traveling salesman: mathematics at the limits
  of computation}.
\newblock Princeton University Press.

\bibitem[\protect\citeauthoryear{Dai \bgroup et al\mbox.\egroup
  }{2017}]{dai2017learning}
Dai, H.; Khalil, E.; Zhang, Y.; Dilkina, B.; and Song, L.
\newblock 2017.
\newblock Learning combinatorial optimization algorithms over graphs.
\newblock In {\em Advances in Neural Information Processing Systems},
  6348--6358.

\bibitem[\protect\citeauthoryear{D{\'o}sa and Sgall}{2013}]{dosa2013first}
D{\'o}sa, G., and Sgall, J.
\newblock 2013.
\newblock First fit bin packing: A tight analysis.
\newblock In {\em 30th International Symposium on Theoretical Aspects of
  Computer Science (STACS 2013)}.
\newblock Schloss Dagstuhl-Leibniz-Zentrum fuer Informatik.

\bibitem[\protect\citeauthoryear{Farahani and
  Hekmatfar}{2009}]{farahani2009facility}
Farahani, R.~Z., and Hekmatfar, M.
\newblock 2009.
\newblock {\em Facility location: concepts, models, algorithms and case
  studies}.
\newblock Springer.

\bibitem[\protect\citeauthoryear{Fischetti and Lodi}{2003}]{fischetti2003local}
Fischetti, M., and Lodi, A.
\newblock 2003.
\newblock Local branching.
\newblock {\em Mathematical programming} 98(1-3):23--47.

\bibitem[\protect\citeauthoryear{Fischetti and
  Monaci}{2014}]{fischetti2014proximity}
Fischetti, M., and Monaci, M.
\newblock 2014.
\newblock Proximity search for 0-1 mixed-integer convex programming.
\newblock {\em Journal of Heuristics} 20(6):709--731.

\bibitem[\protect\citeauthoryear{Gasse \bgroup et al\mbox.\egroup
  }{2019}]{gasse2019exact}
Gasse, M.; Ch{\'e}telat, D.; Ferroni, N.; Charlin, L.; and Lodi, A.
\newblock 2019.
\newblock Exact combinatorial optimization with graph convolutional neural
  networks.
\newblock {\em arXiv preprint arXiv:1906.01629}.

\bibitem[\protect\citeauthoryear{Gopalakrishnan and
  Johnson}{2005}]{gopalakrishnan2005airline}
Gopalakrishnan, B., and Johnson, E.~L.
\newblock 2005.
\newblock Airline crew scheduling: state-of-the-art.
\newblock {\em Annals of Operations Research} 140(1):305--337.

\bibitem[\protect\citeauthoryear{He, Daume~III, and
  Eisner}{2014}]{he2014learning}
He, H.; Daume~III, H.; and Eisner, J.~M.
\newblock 2014.
\newblock Learning to search in branch and bound algorithms.
\newblock In {\em Advances in neural information processing systems},
  3293--3301.

\bibitem[\protect\citeauthoryear{Khalil \bgroup et al\mbox.\egroup
  }{2016}]{khalil2016learning}
Khalil, E.~B.; Le~Bodic, P.; Song, L.; Nemhauser, G.~L.; and Dilkina, B.~N.
\newblock 2016.
\newblock Learning to branch in mixed integer programming.
\newblock In {\em AAAI},  724--731.

\bibitem[\protect\citeauthoryear{Khalil \bgroup et al\mbox.\egroup
  }{2017}]{khalil2017learning}
Khalil, E.~B.; Dilkina, B.; Nemhauser, G.~L.; Ahmed, S.; and Shao, Y.
\newblock 2017.
\newblock Learning to run heuristics in tree search.
\newblock In {\em 26th International Joint Conference on Artificial
  Intelligence (IJCAI)}.

\bibitem[\protect\citeauthoryear{Kool and Welling}{2018}]{kool2018attention}
Kool, W., and Welling, M.
\newblock 2018.
\newblock Attention solves your tsp.
\newblock {\em arXiv preprint arXiv:1803.08475}.

\bibitem[\protect\citeauthoryear{Kool, van Hoof, and
  Welling}{2018}]{kool2018attention2}
Kool, W.; van Hoof, H.; and Welling, M.
\newblock 2018.
\newblock Attention, learn to solve routing problems!
\newblock {\em arXiv preprint arXiv:1803.08475}.

\bibitem[\protect\citeauthoryear{Kruber, L{\"u}bbecke, and
  Parmentier}{2017}]{kruber2017learning}
Kruber, M.; L{\"u}bbecke, M.~E.; and Parmentier, A.
\newblock 2017.
\newblock Learning when to use a decomposition.
\newblock In {\em International Conference on AI and OR Techniques in
  Constraint Programming for Combinatorial Optimization Problems},  202--210.
\newblock Springer.

\bibitem[\protect\citeauthoryear{Laporte}{2009}]{laporte2009fifty}
Laporte, G.
\newblock 2009.
\newblock Fifty years of vehicle routing.
\newblock {\em Transportation Science} 43(4):408--416.

\bibitem[\protect\citeauthoryear{Li, Chen, and
  Koltun}{2018}]{li2018combinatorial}
Li, Z.; Chen, Q.; and Koltun, V.
\newblock 2018.
\newblock Combinatorial optimization with graph convolutional networks and
  guided tree search.
\newblock In {\em Advances in Neural Information Processing Systems},
  537--546.

\bibitem[\protect\citeauthoryear{miplib2017}{2018}]{miplib2017}
2018.
\newblock {MIPLIB} 2017.
\newblock http://miplib.zib.de.

\bibitem[\protect\citeauthoryear{Nazari \bgroup et al\mbox.\egroup
  }{2018}]{nazari2018reinforcement}
Nazari, M.; Oroojlooy, A.; Snyder, L.; and Tak{\'a}c, M.
\newblock 2018.
\newblock Reinforcement learning for solving the vehicle routing problem.
\newblock In {\em Advances in Neural Information Processing Systems},
  9839--9849.

\bibitem[\protect\citeauthoryear{Pinedo}{2012}]{pinedo2012scheduling}
Pinedo, M.
\newblock 2012.
\newblock {\em Scheduling}, volume~29.
\newblock Springer.

\bibitem[\protect\citeauthoryear{Selsam \bgroup et al\mbox.\egroup
  }{2018}]{selsam2018learning}
Selsam, D.; Lamm, M.; B{\"u}nz, B.; Liang, P.; de~Moura, L.; and Dill, D.~L.
\newblock 2018.
\newblock Learning a sat solver from single-bit supervision.
\newblock {\em arXiv preprint arXiv:1802.03685}.

\bibitem[\protect\citeauthoryear{Tang, Agrawal, and
  Faenza}{2019}]{tang2019reinforcement}
Tang, Y.; Agrawal, S.; and Faenza, Y.
\newblock 2019.
\newblock Reinforcement learning for integer programming: Learning to cut.
\newblock {\em arXiv preprint arXiv:1906.04859}.

\bibitem[\protect\citeauthoryear{Veli{\v{c}}kovi{\'c} \bgroup et
  al\mbox.\egroup }{2017}]{velivckovic2017graph}
Veli{\v{c}}kovi{\'c}, P.; Cucurull, G.; Casanova, A.; Romero, A.; Lio, P.; and
  Bengio, Y.
\newblock 2017.
\newblock Graph attention networks.
\newblock {\em arXiv preprint arXiv:1710.10903}.

\bibitem[\protect\citeauthoryear{Vinyals, Fortunato, and
  Jaitly}{2015}]{vinyals2015pointer}
Vinyals, O.; Fortunato, M.; and Jaitly, N.
\newblock 2015.
\newblock Pointer networks.
\newblock In {\em Advances in Neural Information Processing Systems},
  2692--2700.

\bibitem[\protect\citeauthoryear{Zhu}{2004}]{zhu2004recall}
Zhu, M.
\newblock 2004.
\newblock Recall, precision and average precision.
\newblock {\em Department of Statistics and Actuarial Science, University of
  Waterloo, Waterloo} 2:30.

\end{thebibliography}
\bibliographystyle{aaai}

\newpage

\section*{Appendix A}
In this appendix, we describe in table 6 the variable node features, constraint node features, and edge features in detail. All the features are collected at the root node of the B\&B search tree where presolving and root LP relaxation has completed.

\begin{table*}[htbp!]
\footnotesize
\center
\begin{tabular}{@{}llcl@{}}
\toprule
{\textbf{Variable Features}} & Feature Description&    count&  \\ \midrule
\hline
Basic        & variable type (is binary, general integer)  & 2  &  \\
                  & objective function coefficents (original, positive, negative)  & 3  & \\
                  & number of non-zero coefficient in the constraint. & 1  &  \\
		   & number of up (down) locks. & 2  &  \\
                  \hline 
LP              & LP value ($x_j, x_j -\lfloor x_j \rfloor, \lceil x_j \rceil - x_j$) & 3  &\\
		   & LP value is fractional & 1  &  \\
                  & pseudocosts (upwards, downwards, ratio, sum, product)  & 5  &  \\
                  & global lower (upper) bound  &   2 &  \\
                  & reduced cost  &  1  &  \\
                  \hline
Structure    
                  & \tabincell{l}{degree statistics (mean, stdev, min, max) of constraints that the\\ variable has nonzero coefficients.} & 4  &  \\
                  & maximum (minimum) ratio of positive (negative) LHS (RHS) value.  & 8  &  \\
                  & \tabincell{l}{positive (negative) coefficient statistics (count, mean, stdev, \\min, max)  of variables in the constraints. } & 10 &  \\ 
                  & \tabincell{l}{coefficient statistics of variables in the constraints (sum, mean, stdev,\\ max, min) with respect to three weighting schemes: unit weight, dual \\ cost, inverse of the coefficients sum in the constraint.} &  15   & \\
\hline\hline
\textbf{Constraint Features} & &    &  \\ 
\hline
Basic           &\tabincell{l}{constraint type (is singleton, aggregation, precedence, knapsack, logicor, \\ general linear,  AND, OR, XOR, linking, cardinality, variable bound)}&12\\
                   & Left-hand-side (LHS) and right-hand-side (RHS) & 2\\
                   & number of nonzero (positive, negative) entries in the constraint & 3\\
                   \hline
LP               & dual solution of the constraint & 1 \\
                   & basis status of the constraint in the LP solution & 1\\
\hline
Structure      & sum norm of absolute (positive, negative) values of coefficients & 3\\
                   & variable coefficient statistics in the constraint (mean, stdev, min, max) & 4\\
\hline\hline
\textbf{Edge Features}& & &\\
\hline
Basic           &  original edge coefficient & 1 \\
                  & normalized edge coefficient & 1 \\
\bottomrule   
\end{tabular}
\\ \caption{\label{table_feature}Description of variable, constraint and edge features.}
\end{table*}

\section*{Appendix B}
In this appendix, we describe the MIP formulation for each of the 8 types of CO problems in the paper. Table 7 and 8 summarize the number of variables, constraints, percentage of nonzeros in the coefficient matrix $\boldsymbol{A}$, and the percentage of binary variables that takes value 1 in the optimal solution.

\begin{table*}[htbp!]
\footnotesize
\centering
\caption{Problem scale statistics for large-scale instances}
  \begin{tabular}{@{}lcccc@{}}
  \toprule
       & \begin{tabular}[c]{@{}l@{}}Num. of \\ variables\end{tabular} & \begin{tabular}[c]{@{}l@{}}Num. of \\ constraints\end{tabular} & \begin{tabular}[c]{@{}l@{}}Fraction of nonzeros in\\ the coefficient matrix\end{tabular} & \begin{tabular}[c]{@{}l@{}}Fraction of nonzeros in\\ a feasible solution\end{tabular} \\ \midrule
        
        FCNF & [2398, 3568] & [1402, 2078] & 0.00118 & 0.02913\\
        CFL & [28956, 28956] & [29336, 29336] & 0.00014 & 0.01358\\
        GA & [22400, 22400] & [600, 600] & 0.00333 & 0.02500\\
        MIS & [400, 400] & [19153, 19713] & 0.00502 & 0.05815\\
        MK & [765, 842] & [46, 51] & 1.00000 & 0.02684\\
        SC & [4500, 4500] & [3500, 3500] & 0.03000 & 0.02602\\
        TSP & [17689, 19600] & [17954, 19879] & 0.00031 & 0.00737\\
        VRP & [1764, 1764] & [1802, 1802] & 0.00314 & 0.02963 \\
        \bottomrule
      \end{tabular}
  \end{table*}

\begin{table*}[htbp!]
\footnotesize
\centering
\caption{Problem scale statistics for small-scale instances}
    \begin{tabular}{@{}lcccc@{}}
      \toprule
       & \begin{tabular}[c]{@{}l@{}}Num. of \\ variables\end{tabular} & \begin{tabular}[c]{@{}l@{}}Num. of \\ constraints\end{tabular} & \begin{tabular}[c]{@{}l@{}}Fraction of nonzeros in\\ the coefficient matrix\end{tabular} & \begin{tabular}[c]{@{}l@{}}Fraction of nonzeros in\\ a feasible solution\end{tabular} \\ \midrule
          
          FCNF & [1702, 2640] & [996, 1533] & 0.00163 & 0.03392\\
          CFL & [1212, 1212] & [1312, 1312] & 0.00303 & 0.08624\\
          GA & [1152, 1152] & [108, 108] & 0.01852 & 0.08333\\
          MIS & [125, 125] & [1734, 1929] & 0.01620 & 0.14572\\
          MK & [315, 350] & [19, 21] & 1.00000 & 0.07113\\
          SC & [750, 750] & [550, 550] & 0.05000 & 0.06533\\
          TSP & [1296, 1600] & [1367, 1679] & 0.00387 & 0.02697\\
          VRP & [196, 196] & [206, 206] & 0.02422 & 0.08069\\
          \bottomrule
        \end{tabular}
    \end{table*}

\subsection{1) Fixed Charge Network Flow:}

Consider a directed graph $\mathcal{G}=(\mathcal{V}, \mathcal{E})$, where each node $v\in\mathcal{V}$ has demand $d_v$ and the demand is balanced in the graph: $\sum_{v\in\mathcal{V}}=0$. The capacity of an arc $e\in\mathcal{E}$ is $u_e>0$ and the cost of an $x_e>0$ quantity flow on this arc has a cost $f_e + c_e x_e$.
\\
\\
\textbf{Decision variables}:
\begin{itemize}
\item $y_e$: binary variable. $y_e=1$, if arc $e\in\mathcal{E}$ is used and $y_e=0$ otherwise.
\item $x_e$: continuous variable, flow quantity on arc $e\in\mathcal{E}$.
\end{itemize}
\textbf{Formulation}:
\begin{align}
\min~~~ &\sum_{e\in \mathcal{E}}(f_e y_e + c_e x_e)\\
\text{s.t.}~~~  & \sum_{e\in\mathcal{E}(\mathcal{V}, v)} - \sum_{e\in\mathcal{E}(v,\mathcal{V})} = d_v,~~\forall v\in\mathcal{V},\\
&0\le x_e \le u_e y_e, ~~\forall e\in\mathcal{E},\\
&y_e\in\{0,1\}, ~~\forall e\in\mathcal{E}.
\end{align}

\subsection{2) Capacitated Facility Location:}

Suppose there are $m$ facilities and $n$ customers and we wish to satisfy the customers demand at minimum cost. Let $f_i$ denote the fixed cost of building facility $i\in\{1, \ldots, m\}$ and $c_{ij}$ the shipping cost of products from facility $i$ to customer $j\in\{1, \ldots, n\}$. The demand of customer $j$ is assumed to be $d_j>0$ and the capacity of facility $i$ is assumed to be $u_i>0$.
\\
\\
\textbf{Decision variables}:
\begin{itemize}
\item $x_j$: binary variable. $x_j=1$ if facility $j$ is built, and $x_j=0$ otherwise.
\item $y_{ij}$: continuous variable, the fraction of the demand $d_j$ fulfilled by facility $i$.
\end{itemize}
\textbf{Formulation}:
\begin{align}
\min~~~ &\sum_{i=1}^{m}\sum_{j=1}^{n} c_{ij}y_{ij}+\sum_{i=1}^{m}f_i x_i\\
\text{s.t.}~~~  & \sum_{i=1}^{m} y_{ij} = 1,~~\forall j\in\{1, \ldots, n\},\\
&\sum_{j=1}^{n} d_i y_{ij}\le u_i x_i, ~~\forall i \in \{1, \ldots, m\}\\
&y_{ij}\ge 0, ~~\forall i \in \{1, \ldots, m\}, ~j\in\{1, \ldots, n\}\\
&x_j\in\{0,1\}, ~~~\forall j\in\{1, \ldots, n\}.
\end{align}

\subsection{3) Generalized Assignment:}

Suppose there are $n$ tasks and $m$ agents and we wish to assign the tasks to agents to maximize total revenue. Let $p_{ij}$ denote the revenue of assigning task $j$ to agent $i$ and $w_{ij}$ denote the resource consumed, $i \in \{1, \ldots, m\}, j \in \{1, \ldots, n\}$.  The total resource of agent $i$ is assumed to be $t_i$.
\\
\\
\textbf{Decision variables}:
\begin{itemize}
\item $x_{ij}$: binary variable. $x_{ij}=1$ if task  $j$ is assigned to agent $i$, and $x_{ij}=0$ otherwise.
\end{itemize}
\textbf{Formulation}:
\begin{align}
\max~~~ &\sum_{i=1}^{m}\sum_{j=1}^{n} p_{ij}x_{ij}\\
\text{s.t.}~~~  & \sum_{j=1}^{n} w_{ij}x_{ij} \le t_i,~~\forall i\in\{1, \ldots, m\},\\
&\sum_{i=1}^{m} x_{ij}= 1, ~~\forall j \in \{1, \ldots, n\},\\
&x_{ij}\in \{0,1\}, ~~\forall i \in \{1, \ldots, m\}, ~j\in\{1, \ldots, n\}.
\end{align}

\subsection{4) Maximal Independent Set:}
Consider an undirected graph $\mathcal{G}=(\mathcal{V}, \mathcal{E})$, a subset of nodes $\mathcal{S}\in\mathcal{V}$ is called an independent set iff there is no edge between any pair of nodes in $\mathcal{S}$. The maximal independent set problem is to find an independent set in $\mathcal{G}$ of maximum cardinality. 
\\
\\
\textbf{Decision variables}:
\begin{itemize}
\item $x_{v}$: binary variable. $x_{v}=1$ if node $v\in\mathcal{V}$ is is chosen in the independent set, and 0 otherwise.
\end{itemize}
\textbf{Formulation}:
\begin{align}
\max~~~ &\sum_{v\in\mathcal{V}}x_v\\
\text{s.t.}~~~  & x_u + x_v \le 1,~~\forall (u,v)\in \mathcal{E},\\
& x_v\in \{0,1\},~~\forall v\in \mathcal{V}.
\end{align}

\subsection{5) Multidimensional Knapsack:}
Consider a knapsack problem of $n$ items with $m$ dimensional capacity constraints. Let $W_i$ denotes the capacity of the $i$-th dimension in the knapsack, $p_j$ the profit of packing item $j$, and $w_{ij}$ the size of item $j$ in the $i$-th dimension, $i\in\{1,\ldots, m\}, j\in\{1,\ldots,n\}$. 
\\
\\
\textbf{Decision variables}:
\begin{itemize}
\item $x_{j}$: binary variable. $x_{j}=1$ if item $j$ is chosen in the knapsack, and 0 otherwise.
\end{itemize}
\textbf{Formulation}:
\begin{align}
\max~~~ &\sum_{j=1}^n p_j x_j\\
\text{s.t.}~~~  & \sum_{j=1}^{n}w_{ij} x_j \le W_i,~~\forall i\in\{1,\ldots,m\},\\
& x_j\in \{0,1\},~~\forall j\in \{1,\ldots,n\}.
\end{align}

\subsection{6) Set Covering:}
Given a finite set $\mathcal{U}$ and a collection of $n$ subsets  $\mathcal{S}_1,\ldots, \mathcal{S}_n$ of $\mathcal{U}$, the set covering problem is to identify the fewest sets of which the union is $\mathcal{U}$.
\\
\\
\textbf{Decision variables}:
\begin{itemize}
\item $x_{j}$: binary variable. $x_{j}=1$ if set $j$ is chosen, and 0 otherwise.
\end{itemize}
\textbf{Formulation}:
\begin{align}
\min~~~ &\sum_{j=1}^n x_j\\
\text{s.t.}~~~  & \sum_{j\in \{1,\ldots,n\}|v\in\mathcal{S}_j}x_j \ge 1,~~\forall v\in\mathcal{U},\\
& x_j\in \{0,1\},~~\forall j\in \{1,\ldots,n\}.
\end{align}

\subsection{7) Traveling Salesman Problem}
Given a list of $n$ cities, the traveling salesman problem is to find a shortest route to visit each city and returns to the origin city. Let $c_{ij}$ denotes the distance from city $i$ to city $j$ ($i\neq j,~i,j\in\{1,\ldots,n\}$). We use the well-known Miller-Tucker-Zemlin (MTZ) formulation to model the TSP.
\\
\\
\textbf{Decision variables}:
\begin{itemize}
\item $x_{ij}$: binary variable. $x_{ij}=1$ city $j$ is visited immediately after city $i$, and 0 otherwise.
\item $u_j$: continuous variable, indicating the that city $j$ is the $u_j$-th visited city.
\end{itemize}
\textbf{Formulation}:
\begin{align}
\min~~~ &\sum_{i=1}^n\sum_{j\neq i, j=1}^{n} c_{ij}x_{ij}\\
\text{s.t.}~~~  & \sum_{i=1, i\neq j}^n x_{ij} = 1,~~\forall j\in\{1,\ldots,n\},\\
& \sum_{j=1, j\neq i}^n x_{ij} = 1,~~\forall i\in\{1,\ldots,n\},\\
& u_i -u_j+n x_{ij}\le n-1,~~\forall 2\le i\neq j\le n,\\
& 0\le u_i \le n-1, ~~\forall i\in\{2,\ldots, n\},\\
& x_{ij}\in \{0,1\},~~\forall i,j\in \{1,\ldots,n\}.
\end{align}

\subsection{8) Vehicle Routing Problem}
Given a set of $n$ customers, the vehicle routing problem is to find the optimal set of routes to traverse in order to fulfill the demand of customers. To serve the customers, a fleet of $K$ vehicles, each of which has a maximum capacity $Q$ are provided. Let $c_{ij}$ denotes the distance from customer $i$ to customer $j$ ($i\neq j,~i,j\in\{0,\ldots,n+1\}$).
\\
\\
\textbf{Decision variables}:
\begin{itemize}
\item $x_{ij}$: binary variable. $x_{ij}=1$ city $j$ is visited immediately after city $i$ by some vehicle, and 0 otherwise.
\item $y_j$: continuous variable, the cumulated demand on the route that visits node $j$ up to this visit.
\end{itemize}
\textbf{Formulation}:
\begin{align}
\min~~~ &\sum_{i=0}^{n+1}\sum_{j=0}^{n+1} c_{ij}x_{ij}\\
\text{s.t.}~~~  & \sum_{j=1, j\neq i}^{n+1} x_{ij} = 1,~~\forall i\in\{1,\ldots,n\},\\
& \sum_{i=0, i\neq h}^{n} x_{ih} - \sum_{j=1, j\neq h}^{n+1} x_{hj} = 0,~~\forall h\in\{1,\ldots,n\},\\
& \sum_{j=1}^n x_{0j} \le K,\\
& y_j\ge y_i+q_j x_{ij} - Q(1-x_{ij}),~~\forall i,j\in\{0,\ldots,n+1\}\\
& 0 \le y_i \le Q, ~~\forall i\in\{0,\ldots, n+1\},\\
& x_{ij}\in \{0,1\},~~\forall i,j\in \{0,\ldots,n+1\}.
\end{align}

\end{document}